\documentclass{article} 
\usepackage{iclr2026_conference,times}

\usepackage{amsmath,amsfonts,bm}

\def\eqref#1{equation~\ref{#1}}

\def\1{\bm{1}}

\DeclareMathAlphabet{\mathsfit}{\encodingdefault}{\sfdefault}{m}{sl}
\SetMathAlphabet{\mathsfit}{bold}{\encodingdefault}{\sfdefault}{bx}{n}

\usepackage{amsmath}

\usepackage{array,hyperref}
\usepackage[utf8]{inputenc} 
\usepackage{textcomp} 
\usepackage{url,cleveref}
\usepackage{xcolor} 
\usepackage{soul} 
\usepackage{graphicx}
\usepackage{amssymb,bm,enumitem}
\usepackage{arydshln}

\title{Want to train KANS at scale? Now UKAN!}
\author{Alireza Moradzadeh, Srimukh Prasad Veccham, Lukasz Wawrzyniak, Miles Macklin \& Saee G. Paliwal \\
NVIDIA\\
\texttt{\{amoradzadeh,saeep\}@nvidia.com} \\
}

\iclrfinalcopy 
\begin{document}

\maketitle

\begin{abstract}
Kolmogorov–Arnold Networks (KANs) have recently emerged as a powerful alternative to traditional multilayer perceptrons. However, their reliance on predefined, bounded grids restricts their ability to approximate functions on unbounded domains. To address this, we present Unbounded Kolmogorov–Arnold Networks (UKANs), a method that removes the need for bounded grids in traditional Kolmogorov–Arnold Networks (KANs). The key innovation of this method is a coefficient-generator (CG) model that produces, on the fly, only the B-spline coefficients required locally on an unbounded symmetric grid. UKANs couple multilayer perceptrons with KANs by feeding the positional encoding of grid groups into the CG model, enabling function approximation on unbounded domains without requiring data normalization. To reduce the computational cost of both UKANs and KANs, we introduce a GPU-accelerated library that lowers B-spline evaluation complexity by a factor proportional to the grid size, enabling large-scale learning by leveraging efficient memory management, in line with recent software advances such as FlashAttention and FlashFFTConv. Performance benchmarking confirms the superior memory and computational efficiency of our accelerated KAN (warpKAN), and UKANs, showing a $3-30\times$ speed-up and up to $1000\times$ memory reduction compared to vanilla KANs. Experiments on regression, classification, and generative tasks demonstrate the effectiveness of UKANs to match or surpass KAN accuracy. Finally, we use both accelerated KAN and UKAN in a molecular property prediction task,  establishing the feasibility of large-scale end-to-end training with our optimized implementation. 

\end{abstract}

\section{Background}

Neural networks (MLPs) are the workhorse of the current AI and Deep Learning revolution, driving advances in computer vision, language models, computational science, and more recently biology and molecular science \cite{lecun2015deep,goh2017deep,schutt2018schnet, pandey2022transformational,raissi2017physics}. The universal approximation theorem guarantees that MLPs with enough parameters can fit any function. The widespread adoption of MLPs across various disciplines has led to the emergence of exciting applications such as ChatGPT in the large language model (LLM) domain, and AlphaFold in protein structure prediction \cite{vaswani2017attention,jumper2021highly}. However, MLPs suffer from a few drawbacks, particularly generalization for regression tasks.

Recently, the Kolmogorov-Arnold networks (KANs) \citet{kan24} have gained attention as a promising alternative to traditional MLPs, especially in scientific applications, with novel variants currently under development \citet{wkan24,tkan24,abueidda2024deepokandeepoperatornetwork,kiamari2024gkangraphkolmogorovarnoldnetworks,somvanshi2024survey}. The KAN architecture is partially based on the Kolmogorov-Arnold representation theorem \cite{kat61,kat09}, which states that any multivariate function on a bounded domain can be obtained by a finite composition of continuous univariate functions and summation. Mathematically, this can be represented as:

\begin{equation} 
f(x) = \sum_{q=1}^{2n+1} \phi_q \left( \sum_{p=1}^{n} \phi_{q,p}(x_p) \right) 
\label{eq:katheorem}
\end{equation}

where $\phi_{p,q} : [0, 1] \rightarrow \mathbb{R}$ and $\phi_q : \mathbb{R} \rightarrow \mathbb{R}$. Eq. \ref{eq:katheorem} implies that we can approximate any function by summation of univariate functions. A B-spline curve is one of the best methods to parameterize any univariate function by learning the coefficients of the B-spline basis function. 

To extend the KAN architecture beyond what is described by Eq. \ref{eq:katheorem}, \cite{kan24} proposes a new way to build the KAN computational graph with $L$ layers. They assume Eq. \ref{eq:katheorem} is a 2-layer KAN with $n$, $2n$, and 1 nodes. The generalization for $L$ layers starts by defining $n_l$ nodes $\forall l =0, 1, ... ,L$, where $n_l$ is the number of nodes in the $l^{th}$ layer of the computational graph. 

The activation function between node $i$ in layer $l$, ($l$,$i$) and node $j$ in layer $l+1$, ($l+1$, $j$) is denoted by $\phi_{l,j,i}$; the activation value of node $(l+1, j)$ is obtained by summing all incoming post-activation values $\phi_{l,j,i}(x_{l,i})$,

\begin{equation} 
x_{l+1, j} = \sum_{i=1}^{n_l} \phi_{l,j,i}(x_{l,i})
\label{eq:activation}
\end{equation}

In total, there are $n_ln_{l+1}$ activation values and connections between layer $l$ and layer $l+1$. For an input $\mathbf{x}\in \mathbb{R}^{n_0}$, a general KAN network can be written as a composition of $L$ layers,

\begin{equation} 
\operatorname{KAN}(\mathbf{x})
= \bigl(\Phi_{L-1} \circ \Phi_{L-2} \circ \cdots \circ \Phi_{1} \circ \Phi_{0}\bigr)\,\mathbf{x}.
\label{eq:kangeneral}
\end{equation}

where $\Phi_{l}$ is the matrix function of shape $n_{l+1}\times n_l$ with element $(j,i)$ corresponding to $\phi_{l,j,i}$ activation function.  In practice, $\phi(x)$ is the sum of the basis function $b(x)$ (MLP with activation function) and the spline function,

\begin{equation} 
\phi(x) = w_b b(x) + w_s \operatorname{spline}(x)
\label{eq:kanphi}
\end{equation}

where $\operatorname{spline}(x)$ is parameterized as a linear combination of B-splines such that
\begin{equation} 
\operatorname{spline}(x) = \sum_i c_i B_i(x)
\label{eq:kanspline}
\end{equation}

where $c_i$ are learned parameters. A B-spline of order $k+1$ is a collection of piecewise polynomial functions $B_{i, k+1}(t)$ of degree $k$. The locations where these piecewise polynomials connect to each other are known as knots. Given $m+1$ knot values with a uniqueness constraint on $B_{i,k+1}$, we have,

\begin{equation} 
B_{i, k+1}(t) =  \begin{cases}
\text{non-zero}, & \text{if } t_i \le t < t_{i+k+1}\\[4pt]
0, & \text{otherwise }
\end{cases}
\label{eq:bspline}
\end{equation}

The Cox–de Boor formula recursively builds B-spline of order $k$ using,
\begin{equation} 
B_{i,k}(t)
= \frac{t - t_i}{\,t_{i+k} - t_i\,}\, B_{i,k-1}(t)
+ \frac{t_{i+k+1} - t}{\,t_{i+k+1} - t_{i+1}\,}\, B_{i+1,k-1}(t),
\label{eq:coxdeboor}
\end{equation}
where \(B_{i,0}(t)=\mathbf{1}_{[t_i,\,t_{i+1})}(t)\). 

KANs are promising, but practical use is hampered by compute- and memory-inefficient implementations and reliance on periodic grid updates. In naïve GPU code, evaluating B-spline bases via the Cox–de Boor recursion redundantly recomputes along the entire knot vector (effectively repeating work $k+1$ times per evaluation) and inflates both memory traffic and FLOPs. Many implementations also materialize all intermediate basis states, which depresses batch sizes and GPU utilization. Algorithmically, most public code evaluates B-spline bases in time $\mathcal{O}\!\left(k\,d_g\,d_{\mathrm{in}}\,d_{\mathrm{out}}\right)$, where $k$ is the B-spline degree, $d_g$ the number of knots (we use “grid” for the sorted knot vector), and $d_{\mathrm{in}}(=n_l)$, $d_{\mathrm{out}}(=n_{l+1})$ the input and output dimensions. Training can also stall when inputs drift outside the initialized knot support, where the B-spline basis vanishes; while \cite{kan24} advocates periodic grid updates, this procedure is numerically brittle and ill-suited to reliable, high-throughput GPU execution.

This work removes these bottlenecks in the original KAN by enabling true batched evaluation and eliminating unnecessary intermediates, substantially reducing both memory footprint and FLOPs. Because B-splines also underlie pre-KAN learnable activations, the same kernels accelerate those methods as well.\cite{laf20} Our implementation facilitates a fair comparison between MLPs and KANs; correcting prior apples-to-oranges comparisons that pitted highly optimized MLPs against vanilla implementation of KANs\cite{unfair2024}. As recent ML systems work has shown (e.g., FlashAttention/FlashFFTConv \cite{flashatt,flashatt2,flashatt3,flashfftconv}), careful software design unlocks scale; we follow the same philosophy for KAN-style architectures. Finally, we introduce UKAN to eliminate grid updates during training and present experiments comparing MLP, KAN, and UKAN across a range of settings.

\section{Algorithm}
\label{gen_inst}

Our remedy for KAN's compute and memory overhead is to exploit the compact support of B-splines rather than evaluating the Cox-de Boor recurrence over the entire knot vector. A degree-k basis $B_{i,k}$ is nonzero only on $[t_i, t_{i+k+1})$. Leveraging this observation, we represent the B-spline function with basis matrices \cite{qin1998general} as shown in Eq. \ref{eq:matrixform}.

\begin{equation}
\operatorname{spline}(u) = U \mathbf{M} C
\label{eq:matrixform}
\end{equation}

where $u$ is $\frac{x - t_i}{t_{i+1} - t_i}$, where $i = \lfloor{\frac{x - t_0}{h}}\rfloor$ and $h$ is the distance between two adjacent grid points ($h := t_{i+1} - t_i$). $U$ is the vector of $(1, u, u^2, ..., u^k)$. $\mathbf{M}$ is basis matrix and $C$ is the vector of B-spline coefficients $(c_0, c_1, ..., c_k)$. The basis matrices are obtained by applying recursive B-spline equations, and only depends on the degree of the B-spline function (see Appendix \ref{cubic_bspline} for cubic B-spline basis matrix representation).

We provide efficient implementations of the above formula using NVIDIA Warp \cite{warp2022} in a new library called warpKAN with evaluation complexity of $\mathcal{O}(k d_{in} d_{out})$ along with PyTorch bindings \cite{paszke2019pytorch}. This implementation offers both memory and computational efficiency, as described in Table \ref{Complexity}. However, the reduction in compute and memory cost of B-spline components does not solve the issue of the bounded range of the grid in the original KAN.

\begin{table}[t]
\centering
\caption{Compute complexity of Torch- and Warp-KAN for single layer B-spline evaluation.}
\begin{tabular}{ll}
 \hline
 Model & Complexity  \\ 
 \hline
 Torch KAN & $\mathcal{O}(K d_g d_{in} d_{out})$ \\ 
 Warp KAN & $\mathcal{O}(K d_{in} d_{out})$ \\ 
 Warp UKAN & $\mathcal{O}(K d_{in} d_{out}) + \mathcal{O}_{CG}(d_{emb}^2 + d_{emb}d_{out} K)$ \\ 
 \hline
\end{tabular}
\label{Complexity}

\end{table}

\begin{figure}[h]
\centering
\includegraphics[width={5.5cm}]{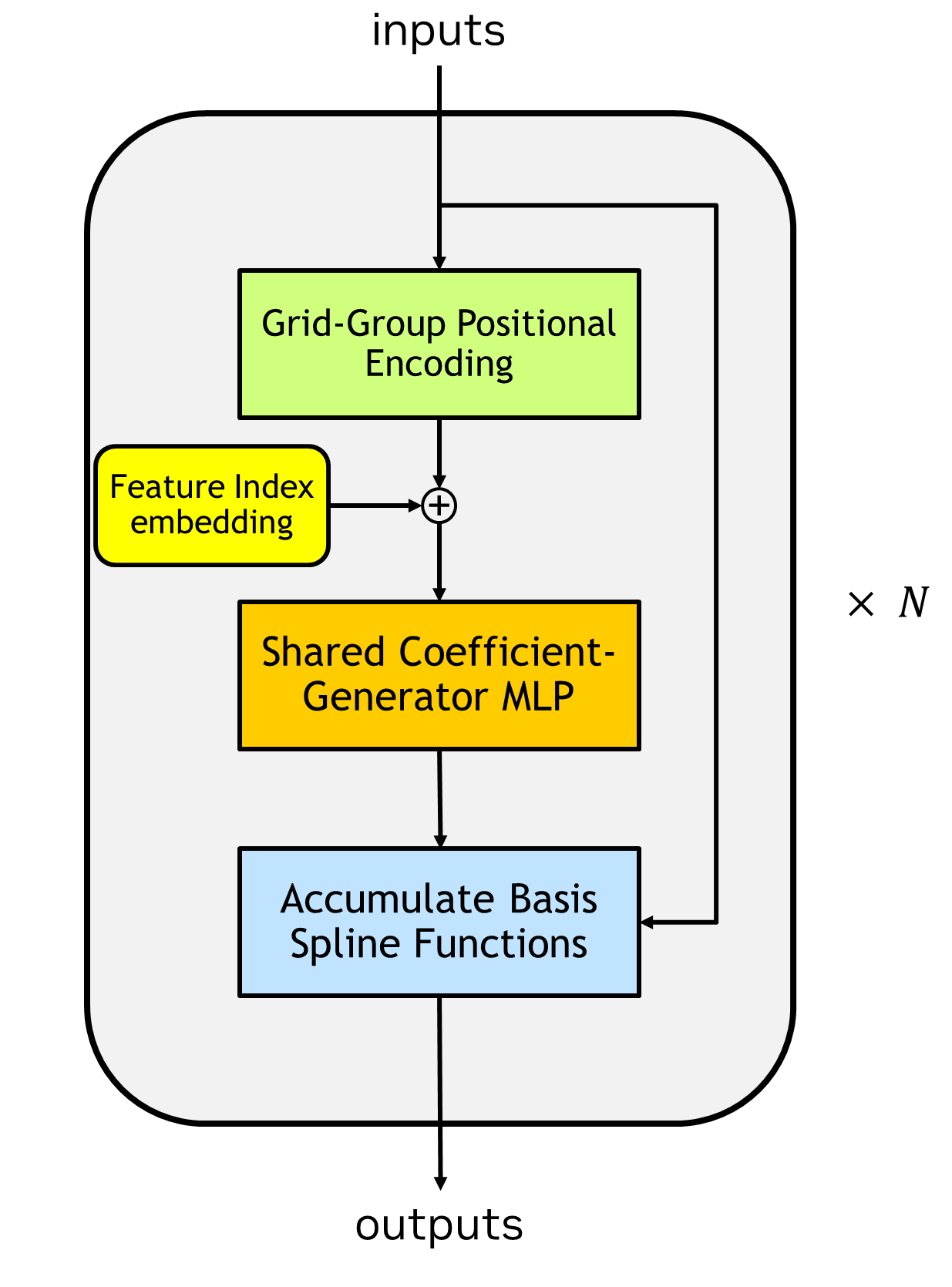}
\caption{The UKAN model architecture including grid group positional encoding, coefficient-generator MLP, and B-spline function.}
\label{fig:ukan}
\end{figure}

We achieve an unbounded grid by generating B-spline coefficients with a coefficient-generator (CG) MLP, akin to Hyena’s filter generation \cite{poli2023hyenahierarchylargerconvolutional}. For a degree-$k$ spline, each evaluation needs $K=k+1$ coefficients (Eq. \ref{eq:matrixform}). A naïve approach—calling the same MLP $K$ times for the $K$ adjacent grid indices performed poorly in our experiments, likely due to missing joint conditioning across those $K$ coefficients. Instead, we partition the uniform grid $\{t_j=j\,h\  : \forall j \in \{..., -2, -1, 0, 1, 2, ...\}\}$ into groups of $K$ consecutive cells. For any $x$, define the cell index $i(x)=\lfloor x/h\rfloor$, the group index

\begin{equation}
g=\Big\lfloor\tfrac{i(x)}{K}\Big\rfloor=\Big\lfloor\tfrac{x}{K\,h}\Big\rfloor,
\end{equation}

and define the within-group offset $r=i(x)\bmod K$. The CG-MLP takes as input the concatenation of (i) an embedding of the feature index and (ii) a sinusoidal positional encoding of the group index $g$ (as in Transformers \cite{vaswani2017attention}), and outputs a vector $C_g\in\mathbb{R}^K$ of coefficients for group $g$. To ensure the correct sliding window across group boundaries, we concatenate the previous and current outputs $[C_{g-1};\,C_g]\in\mathbb{R}^{2K}$ and select the $K$ coefficients

\begin{equation}
\big[C_{g-1};\,C_g\big]_{r:\,r+K-1},
\end{equation}

which are then used in the basis-matrix evaluation of Eq. \ref{eq:matrixform}. This grouping yields an unbounded, index-conditioned parameterization with fewer MLP calls, while preserving the exact $K$-wide local stencil required by B-splines. The general architecture of UKAN is shown in Figure \ref{fig:ukan}. After training, one can materialize all coefficients on any finite interval and store them as KAN parameters, retaining interpretability and enabling the usual symbolic-regression or pruning workflows, which is one of the strong features of KAN architecture (see Appendix \ref{KANUKAN})

\section{Experiments}

In this section, we benchmark the performance of warpKAN for various B-spline orders and grid sizes. Subsequently, we showcase the capabilities of UKAN across a wide-range of tasks commonly encountered in machine learning: classification, regression, approximation, generation, and a real-world application to drug discovery.

\subsection{Performance Benchmarking}

\begin{figure*}[h]
  \centering
   \includegraphics[width=\linewidth]{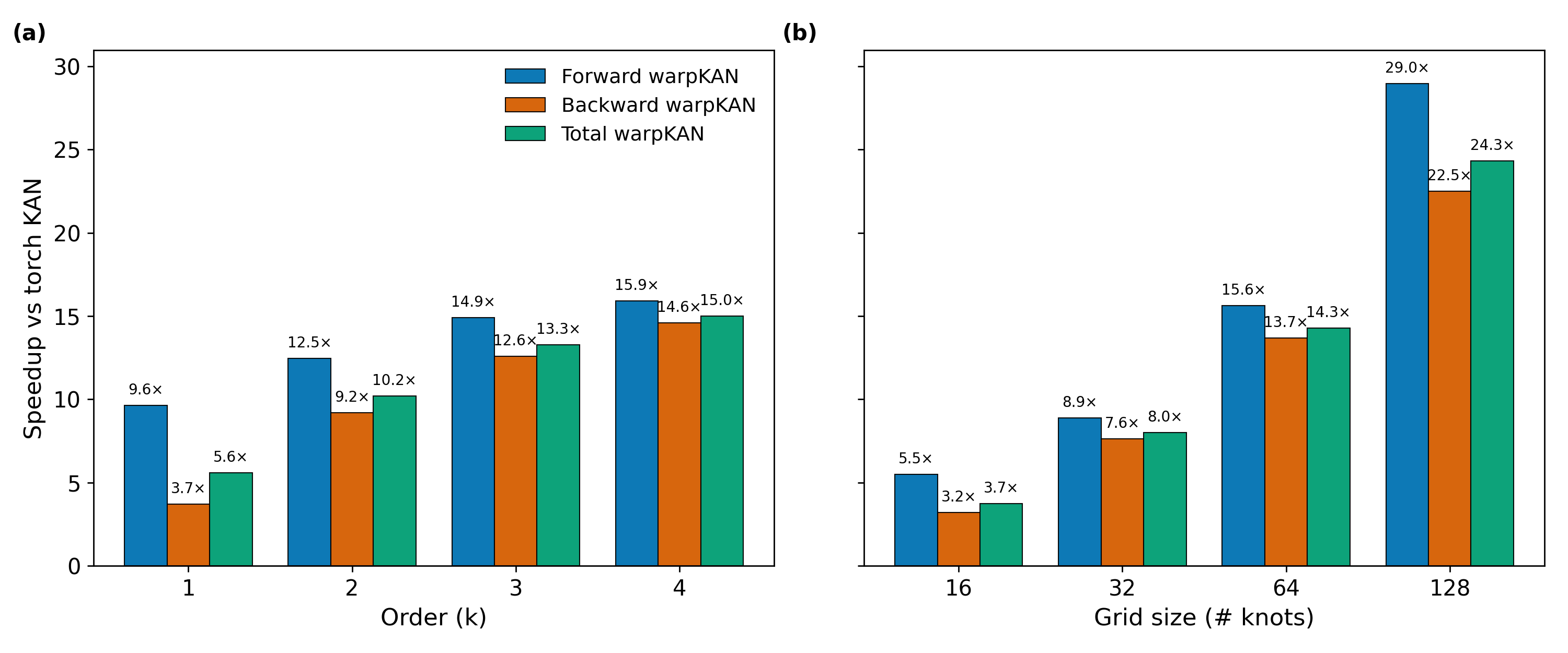}
  \caption{warpKAN vs.\ torchKAN. 
  (a) Increasing B\mbox{-}spline order yields larger speedups ($5.5\!\times$–$15\!\times$). 
  (b) Increasing grid size shows average $\sim\!12\!\times$ and up to $24\!\times$ speedup; 
  torchKAN hits OOM $\ge 256$ knots while warpKAN scales to $2^{18}$. 
  All values are normalized to the public PyTorch implementation.}
  \label{fig:bench}
\end{figure*}

In Figure~\ref{fig:bench}, we benchmark warpKAN against torchKAN under two settings: 
(a) varying the B\mbox{-}spline order and (b) varying the number of knots (“grid size”). 
We report the speedup for the forward and backward passes, as well as their sum, for a single 
layer $\mathrm{KAN}(32\!\to\!32)$. For panel (a), we use a grid size of 64 and a batch size of $2^{16}$; 
for panel (b), we fix the B\mbox{-}spline order to 3 and use a batch size of $2^{17}$. 
All warpKAN results are normalized to the PyTorch implementation from the original paper. 
Across B\mbox{-}spline orders, warpKAN is $5.5\!\times$–$15\!\times$ faster, with speedup increasing with the order. 
Across grid sizes, warpKAN averages about $12\!\times$ and reaches up to $24\!\times$ speedup for larger grids. 
Moreover, torchKAN runs out of memory for grids $\ge 256$, while warpKAN scales to grid sizes up to $2^{18}$; over 
$1000\!\times$ larger. \textbf{Notation:} $\mathrm{Layer}(a\!\to\!b)$ denotes input and output dimensions.

From Figure~\ref{fig:bench}, the naïve PyTorch implementation clearly under-utilizes the GPU. Further gains are expected by (i) exploiting Tensor Cores, (ii) fusing kernels to cut launch overhead and off-chip traffic, and (iii) other common techniques in CUDA programming. These optimizations should shift the workload toward the roofline limits and substantially increase throughput, mirroring the trajectory of the FlashAttention family~\cite{flashatt,flashatt2,flashatt3}. 

We quantify how close our implementation is to the hardware limits using a roofline model. The Speed-of-Light (SoL) time, $t_{\mathrm{SoL}}$, corresponds to the limiting factor between compute (FLOP/s) and memory bandwidth (B/s), and thus represents the best achievable runtime on the hardware. Based on this definition, we construct compute and memory models of the B-spline below.


\begin{table}[h]
\centering
\small
\caption{B-spline forward SoL and runtime vs.\ grid size on A6000 (FP32). $t_{meas}$ is our empirical runtime, the remaining columns are theoretical upper and lower bounds. 
Setup: $N{=}2^{17}$, $d_{\text{in}}{=}d_{\text{out}}{=}32$, order $k{=}3$ ($K{=}4$),
$P_{\mathrm{FLOP}}{=}38.7$ TFLOP/s, $P_{\mathrm{BW}}{=}768$ GB/s.}
\label{tab:bspline_grid_scaling}
\begin{tabular}{rcccccc}
\hline
$G$ & $t_{\text{meas}}$ (ms) & $t_{\text{SoL}}^{\text{best}}$ (ms) & $t_{\text{SoL}}^{\text{worst}}$ (ms)
& $t_{\text{compute}}$ (ms) & $t_{\text{mem}}^{\text{best}}$ (ms) & $t_{\text{mem}}^{\text{worst}}$ (ms) \\
\hline
\phantom{0}32   & 4.9489 & 0.1387 & 2.8399 & 0.1387 & 0.0439 & 2.8399 \\
\phantom{0}64   & 5.1764 & 0.1387 & 2.8399 & 0.1387 & 0.0441 & 2.8399 \\
128  & 5.3669 & 0.1387 & 2.8399 & 0.1387 & 0.0444 & 2.8399 \\
256  & 5.5305 & 0.1387 & 2.8399 & 0.1387 & 0.0451 & 2.8399 \\
512  & 5.7400 & 0.1387 & 2.8399 & 0.1387 & 0.0464 & 2.8399 \\
1024 & 5.9188 & 0.1387 & 2.8399 & 0.1387 & 0.0492 & 2.8399 \\
\hline
\end{tabular}
\label{table:SoL}
\end{table}

\paragraph{Compute model}

Let $N$ be the batch size, $d_{\mathrm{in}}$ and $d_{\mathrm{out}}$ the input/output dims, $k$ the B-spline order,
$K = k{+}1$, and $G$ the number of grid positions (knots) per $(i,o)$ pair.\footnote{Our kernels use a local matrix form; $G$ denotes the global grid size before local selection.}
We write the per-(input,output) multiply–add count as a $k$-dependent constant $c_k$; in our implementation we use
$c_k \approx 2K(K{+}1)$ (the factor 2 accounts for fused multiply–adds counted as two FLOPs).
Then the total work is
\begin{equation}
F_{\mathrm{fwd}} \;\approx\; N \cdot d_{\mathrm{in}} \cdot d_{\mathrm{out}} \cdot c_k.
\label{eq:F_fwd}
\end{equation}

\paragraph{Memory models}
We separate optimistic (best-case) and pessimistic (worst-case) coefficient reuse (no-cache and cached).
We assume contiguous reads of inputs and writes of outputs; coefficients are indexed locally
($K$ per sample) but stored globally ($(G{+}K)$ per $(i,o)$, including padding).

\emph{Best-case reuse} (stream all unique coefficients once across the batch):
\begin{equation}
B_{\mathrm{best}} \;\approx\; s
\Big[ 
N\, (d_{\mathrm{in}} + d_{\mathrm{out}}) \;+\; d_{\mathrm{in}} d_{\mathrm{out}}\, (G{+}K)
\Big],
\qquad s=4~\text{bytes (FP32)}.
\label{eq:B_best}
\end{equation}

\emph{Worst-case reuse} (fetch local $K$ coefficients per sample):
\begin{equation}
B_{\mathrm{worst}} \;\approx\; s
\Big[
N\, (d_{\mathrm{in}} + d_{\mathrm{out}}) \;+\; N\, d_{\mathrm{in}} d_{\mathrm{out}}\, K
\Big].
\label{eq:B_worst}
\end{equation}

The SoL times are then
\begin{equation}
t_{\mathrm{SoL}}^{\mathrm{best}} = \max\!\left(\frac{F_{\mathrm{fwd}}}{P_{\mathrm{FLOP}}}, \frac{B_{\mathrm{best}}}{P_{\mathrm{BW}}}\right),
\qquad
t_{\mathrm{SoL}}^{\mathrm{worst}} = \max\!\left(\frac{F_{\mathrm{fwd}}}{P_{\mathrm{FLOP}}}, \frac{B_{\mathrm{worst}}}{P_{\mathrm{BW}}}\right).
\end{equation}

We perform experiments for different configurations on the A6000 GPU in FP32, and report results in Table\ref{table:SoL}. We time forward calls with CUDA synchronization before and after, discarding warm-up.
We use fixed shapes $(N ,d_{\mathrm{in}} ,d_{\mathrm{out}},k) = (2^{17}, 32, 32, 3)$, report the average per-call time, and compute
$F_{\mathrm{fwd}}$ via~\eqref{eq:F_fwd} with $c_k{=}2K(K{+}1)$.
We report both $t_{\mathrm{SoL}}^{\mathrm{best}}$ and $t_{\mathrm{SoL}}^{\mathrm{worst}}$ using~\eqref{eq:B_best}–\eqref{eq:B_worst}. As shown in Table \ref{table:SoL}, there are still opportunities to improve runtime of B-spline used in KANs and many other workloads. The future work will try to close the gap between runtime and SoL time.\footnote{We do not provide a full roofline breakdown for UKAN, as its cost is dominated by the CG network. Our ongoing work explores a fused implementation that co-schedules the CG MLP with spline evaluation and accumulation in a single kernel; so that coefficients are produced and consumed without intermediate HBM round-trips. These experiments are in progress.} The runtime of vanilla KAN implementations can be estimated by multiplying the speedup values reported in Figure~\ref{fig:bench} with the measured $t_{\mathrm{meas}}$ from Table~\ref{table:SoL}, provided the configuration does not run out of memory.

\subsection{Tasks}

\subsubsection{Regression}
To evaluate the accuracy of our UKAN and KAN in regression, we conducted three experiments.

\begin{enumerate}[label=\Roman*.]
    \item $f(x,y) = \exp ( J_0(20x)+y^2)$, where $J_0$ is the Bessel function of order 0.
    \item $f(x,y) = \exp(\sin \pi x + y^2)$
    \item $f(\mathbf{x}) = \exp(\frac{1}{15}\sum_i^n\sin( {(\frac{4i}{15}+ 1)\pi x_i}) )$	, where $i=0,1,…,15$ and is a high dimensional function compared with functions I and II.
\end{enumerate}

\begin{figure}[h]
    \centering
    \begin{minipage}[b]{0.497\textwidth}
        \centering
        \includegraphics[width=\textwidth]{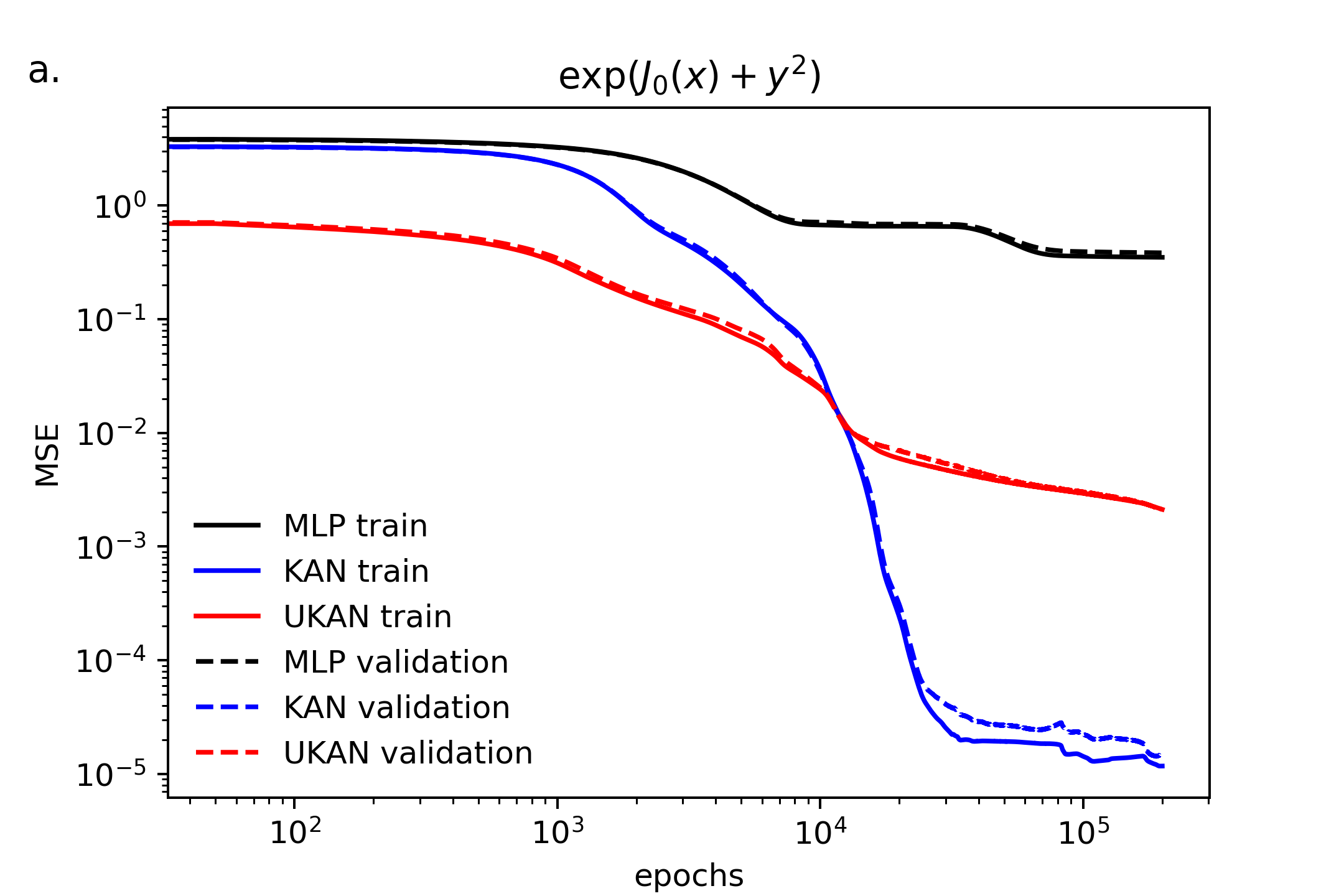}
    \end{minipage}
    \hfill
    \begin{minipage}[b]{0.497\textwidth}
        \centering
        \includegraphics[width=\textwidth]{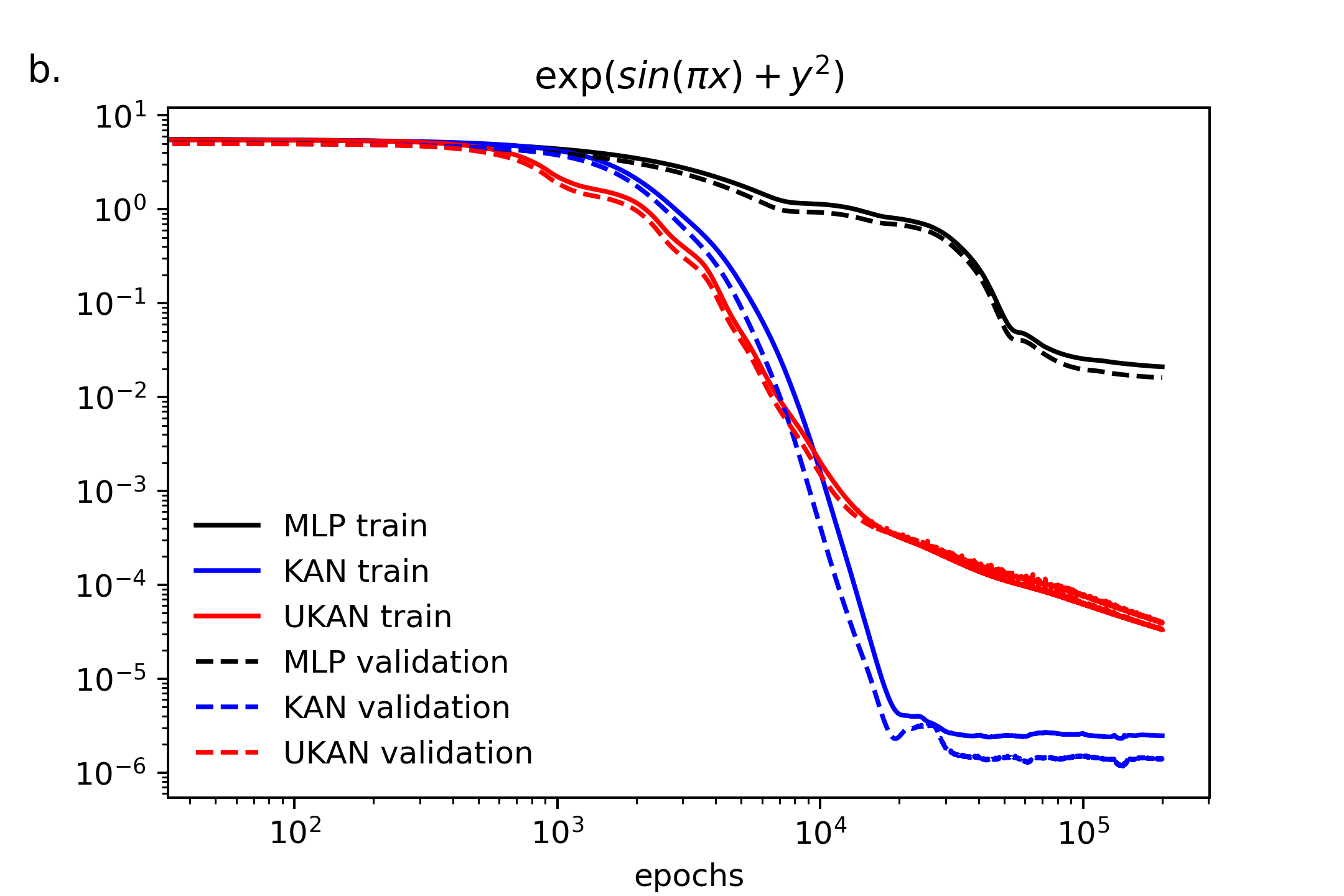}

    \end{minipage}
    \hfill
    \begin{minipage}[b]{0.497\textwidth}
        \centering
        \includegraphics[width=\textwidth]{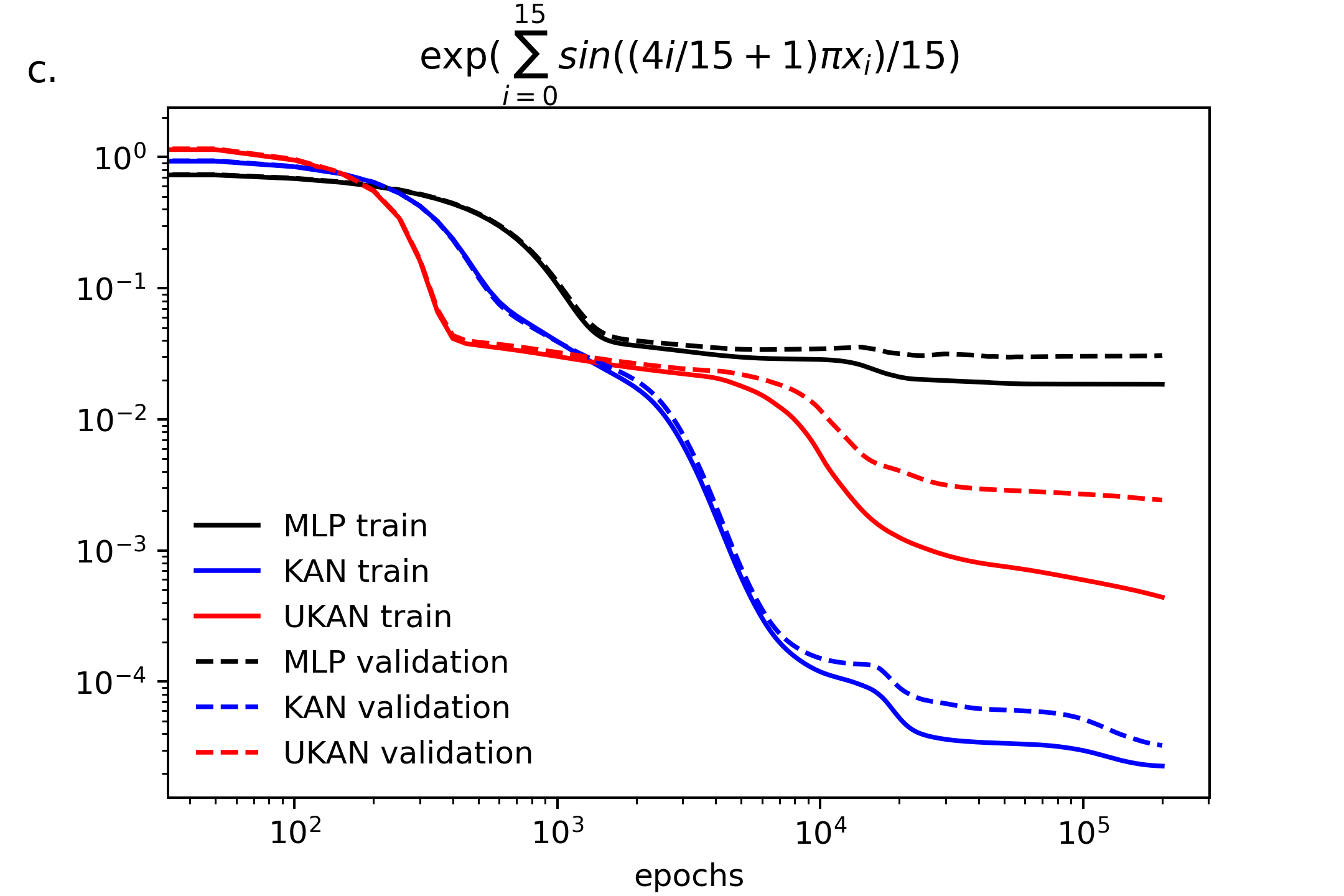}
    \end{minipage}
    \caption{Regression task results. 
(a) RMSE vs.\ training epochs for Function I using KAN, UKAN, and \(\mathrm{MLP}(2\!\to\!5\!\to\!1)\).
(b) RMSE vs.\ training epochs for Function II using KAN, UKAN, and \(\mathrm{MLP}(2\!\to\!5\!\to\!1)\).
(c) RMSE vs.\ training epochs for Function III using KAN, UKAN, and \(\mathrm{MLP}(16\!\to\!32\!\to\!1)\).
}
    \label{fig:three_figures}
\end{figure}

We compare the results of UKAN, KAN and MLP \((2\!\to\!5\to\!1)\) for 2D functions (functions I and II). For function III, we use UKAN, KAN, and MLP \((16\!\to\!32\to\!1)\). For the CG model, we used a two-layer MLP with 8- and 16-dimensional positional encodings and feature embeddings for 2D and 16D functions, respectively. The first layer of the CG MLP uses SiLU nonlinearity and generated coefficients are scaled by another learnable parameter to improve learning, analogous to the original KAN paper. An Adam optimizer \cite{kingma2014adam} with a learning rate of 0.01 and weight decay of $1e^{-5}$ for 200,000 epochs is used to minimize the MSE loss. The learning rate is decayed exponentially with the rate of $1-1e^{-4}$ and minimum learning rate of $1e^{-4}$.  The results are shown in Figure \ref{fig:three_figures}, where UKAN and KAN perform much better than MLP, and KAN performs better than UKAN. In theory, KAN and UKAN have the same learning capacity, but the MLP component of UKAN might slightly hurt generalization and performance compared to KAN. We also note that although the compute cost of a KAN is greater than that of an MLP by a factor of $K$, KAN convergence often compensates for this factor. We also performed a regression task on the n-body problem based on \cite{satorras2022enequivariantgraphneural}, the results are available in Appendix \ref{n-body}.

\subsubsection{Classification}

We devised two experiments to evaluate UKAN's performance in classification tasks. In particular, we trained UKAN and KAN over Moon- and MNIST-datasets.  We report the accuracy of training and validation of both models on the moon-dataset in Table \ref{moondataset}, where we observe close to 100\% accuracy for Moon-dataset with 2D inputs with slight superior accuracy of UKAN over KAN in this setup. Both UKAN and KAN \((2\!\to\!4\to\!2)\) are trained using SGD optimizer with learning rate of 0.01 for 10000 epochs, and results are averaged over 3 different initializations. 

\begin{table}[h]
\centering
\caption{Moon dataset classification accuracy.}
\begin{tabular}{lll}
 \hline
 Model & Training &  Validation \\ 
 \hline
 KAN  & $98.46 \pm 1.3$           & $98.53 \pm 0.4$ \\ 
 UKAN & $\mathbf{100.0 \pm 0.}$ & $\mathbf{99.83 \pm 0.17}$ \\ 
 \hline
\end{tabular}
\label{moondataset}

\end{table}

The final classification task was performed on the MNIST dataset, where we trained both the UKAN and KAN models with configurations \((784\!\to\!32\to\!10)\) and a degree-3 B-spline. Both models were optimized using the Adam optimizer combined with an Exponential scheduler, having a learning rate of $2 \times 10^{-4}$ and a decay rate of $1 - 10^{-4}$. The KAN network incorporated 51 grid points across the interval $[-10, 10]$, whereas UKAN utilized a grid delta of 3.0 and a 48-dimensional positional encoding. Notably, both models employed only the B-spline component without any MLP components. Training was halted upon detection of overfitting in the training dataset. Furthermore, three rounds of independent training with different initializations were conducted to compare the performance of UKAN and KAN. The results, as presented in Table \ref{mnisttable}, indicate that UKAN outperforms KAN on the validation dataset while slightly underperforming on the training dataset.

\begin{table}[h]
\centering
\caption{MNIST dataset classification accuracy.}
\begin{tabular}{lll}
 \hline
 Model & Training &  Validation \\ 
 \hline
 KAN  & $\mathbf{98.93 \pm 0.78}$           & $95.35 \pm 0.04$ \\ 
 UKAN & $98.40 \pm 0.24$ & $\mathbf{96.29 \pm 0.08}$ \\ 
 \hline
\end{tabular}
\label{mnisttable}

\end{table}


\subsubsection{Approximation}
\begin{figure}[h]
\centering
\includegraphics[width={8cm}]{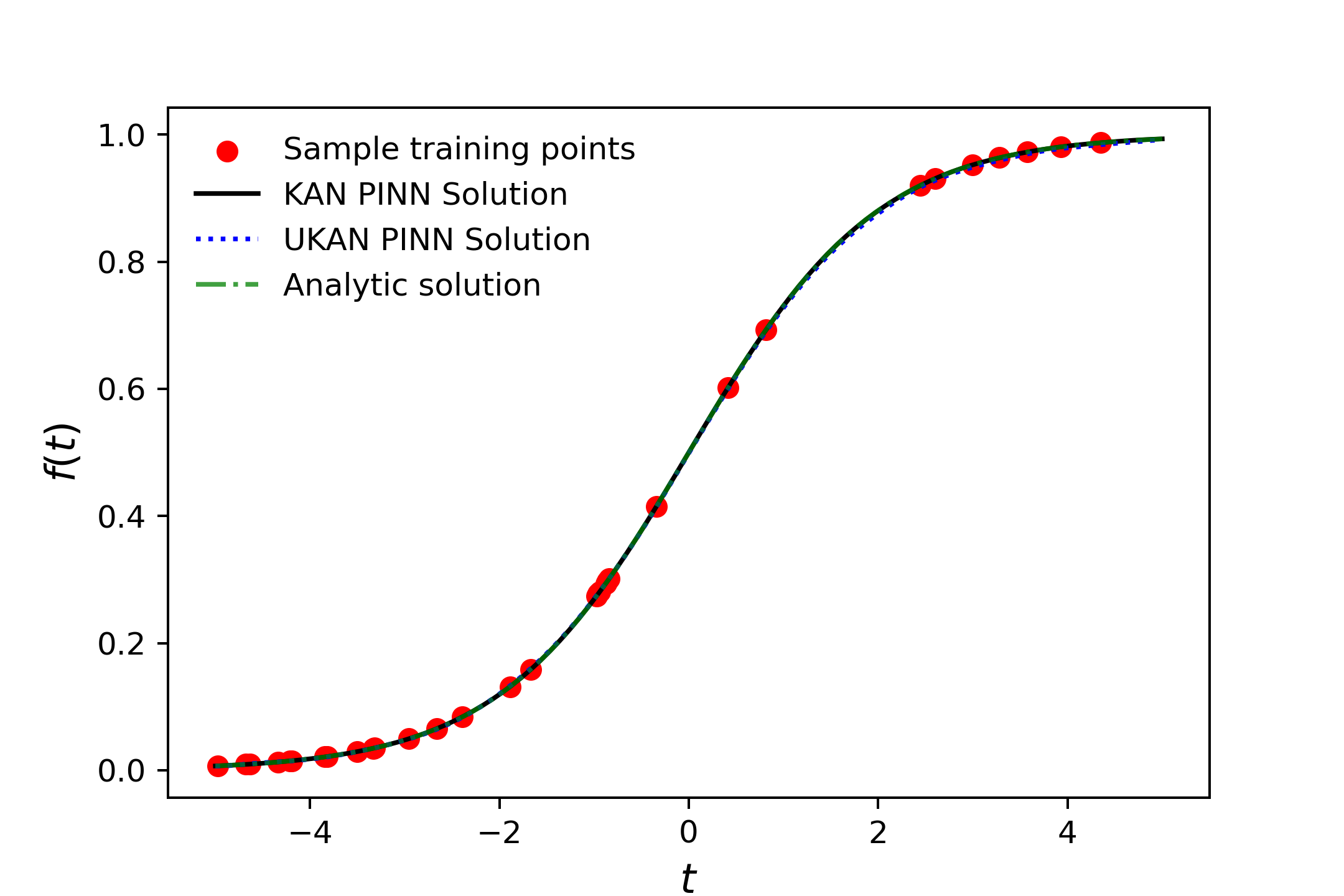}
\caption{KAN and UKAN used in PINNs. Solving logistic growth model with both KAN and UKAN \((1\!\to\!5\to\!1)\) over domain of -5 and 5.}
\label{fig:pinn}
\end{figure}

We explore the effectiveness of UKAN and KAN in physics-informed neural networks \cite{karniadakis2021physics} to solve the logistic growth model, which is used to model population dynamics in biological and ecological systems. For this experiment, we use both UKAN and KAN ($1\rightarrow 5 \rightarrow 1$) without MLP component to solve the differential equation below,
\begin{equation}
\frac{d f}{d t} = R f(t) ( 1 - f(t))
\label{eq:lgm}
\end{equation}

where $R$ is the growth rate set to 1.0 and the function $f(t)$ represents the growth rate of the population over time (t). We impose boundary condition of $f(0)=0.5$ to uniquely specify the solution and compare the results with the analytical solution of $f(t) = \frac{1}{1+\exp{(-t)}}$. We use domain of $[-5,5]$ to sample data and Adam optimizer with the learning of rate of $1e^{-3}$ and weight decay of $1e^{-5}$ and follow the standard procedure for PINN minimization, i.e. minimizing MSE of the differential equation residual over collocation points and boundary conditions. The results are shown in Figure \ref{fig:pinn}, for  and KAN \((1\!\to\!5\to\!1)\) without the MLP component. UKAN and KAN achieve MSE of $1e^{-5}$ and $1e^{-6}$, respectively on the sample dataset, indicating both models are applicable to physics-informed neural networks scenarios.

\subsubsection{Generation}
\begin{figure}[h]
\centering
\includegraphics[width={\textwidth}]{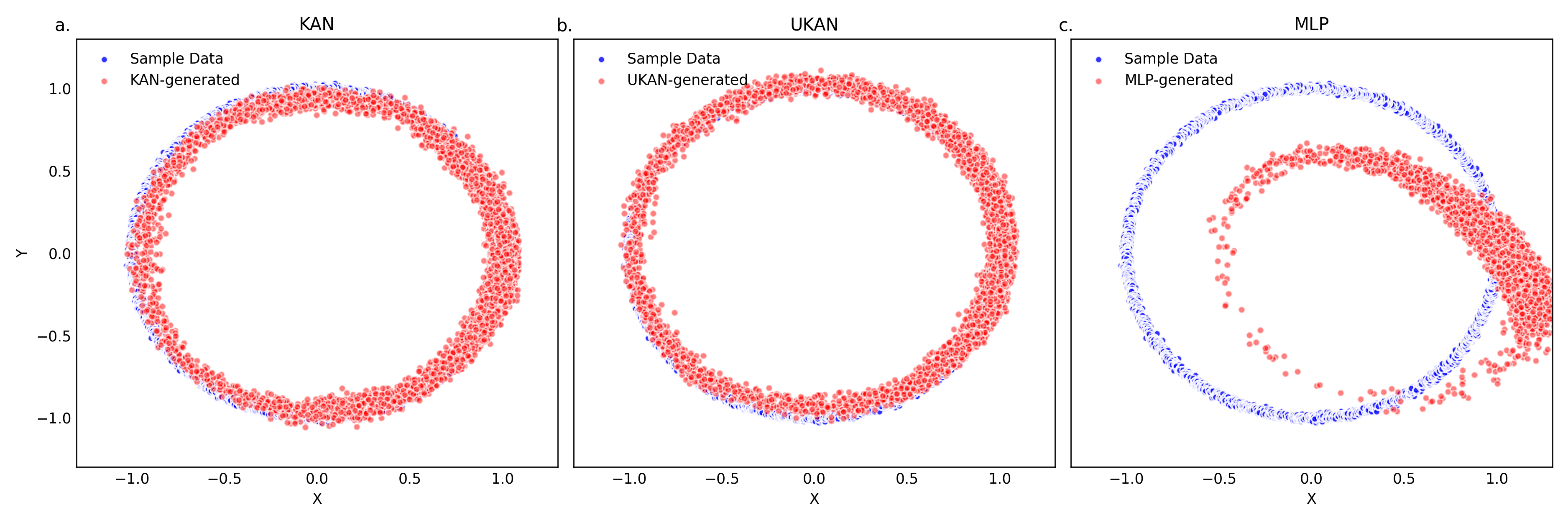}
\caption{DDPM with KAN, UKAN, and MLP.}
\label{fig:ddpm}
\end{figure}

As an example in generative model learning, we evaluated the performance of three different architectures for Denoising Diffusion Probabilistic Models (DDPM) \cite{ddpm2020} on a synthetic 2D circle dataset with added noise. The first architecture is only composed of MLPs, while other architectures use KAN and UKAN in the input of temporal layers and output layer (see Appendix \ref{model_arch} for full architecture). We used Adam optimizer with a learning rate of $5e^{-5}$ for 500 epochs with a batch size of 800. Our results demonstrated that both KAN and UKAN significantly outperformed MLP in terms of the Wasserstein distance shown in Table \ref{ddpm} and sample quality as shown in Figure \ref{fig:ddpm}. Data samples from original distribution and generated from DDPM with KAN, UKAN, and MLP architectures indicates superior performance of KAN and UKAN compared to MLP and slightly superior performance of UKAN over KAN. This result indicates possible applications of KAN and UKAN in generative tasks, where MLP alone might fail to learn underlying data distribution especially in sample quality as we observed loss values of MLP, KAN and UKAN were very small. 

\begin{table}[h]
\centering
\caption{DDPM with KAN, UKAN, and MLP}
\begin{tabular}{|>{\centering\arraybackslash}m{3cm}|>{\centering\arraybackslash}m{3cm}|}
 \hline
 Model & Wasserstein distance\\ 
 \hline
 KAN  &        $      0.693  $  \\ 
 UKAN &        $      \mathbf{0.655}$ \\
 MLP  &        $      1.058$ \\
 \hline
\end{tabular}
\label{ddpm}

\end{table}

\subsubsection{Real-world Application: Drug Discovery}
Accurate \textit{in silico} prediction of molecular properties is key to accelerating drug discovery by early identification of compounds with favorable ADME (Absorption, Distribution, Metabolism, and Excretion) profiles. \cite{ferreira2019admet, beckers2023prediction, seal2025machine} Machine learning methods have shown strong performance in predicting these properties from chemical structure or computed features. The usage of KANs in predicting molecular properties was originally introduced by \cite{li2025kolmogorov}. Due to implementation discrepancies and known issues within the MoleculeNet datasets,\cite{BetterBenchmarks2023Walter} we reimplemented the original model using the warpKAN package and incorporated the necessary corrections. A comprehensive analysis of these results is presented in the Appendix \ref{moleculenet_results_appendix}.

Here, we further demonstrate the ability of UKANs to predict ADME properties, and used a fixed molecular representation to isolate the impact of the KAN and UKAN addition. We choose Morgan fingerprints and RDKit 2D descriptors as our two computed molecular features. In order to overcome the limitations of the MoleculeNet dataset, we use four datasets released by \cite{fang2023prospective}. These models were trained for 200 epochs using Adam optimizer with a learning rate of 1e-4. As shown in Table \ref{table:biogen_adme}, UKANs show superior or comparable performance with respect to KANs for predicting molecular properties from computed features using mean absolute error metric. UKANs outperform KANs on the permeability and solubility dataset using Morgan fingerprint features while their performance is statistically equivalent (using T-test for means) on the other two datasets. Using RDKit 2D descriptors, a set of 200 molecular descriptors describing the 2D structure of the molecule, UKAN shows better performance than KAN on the Microsomal stability (rat) dataset. These results show that UKANs are superior to KANs in accurately predicting molecular properties relevant to the pharmaceutical industry. Details about model architecture and data preprocessing can be found in the Appendix \ref{mol_prop_pred_arch_appendix} and \ref{mol_prop_prep_data_appendix} respectively.

\begin{table}[h]
\centering
\caption{Performance of KAN and UKAN on Biogen ADME datasets using Morgan fingerprint (labeled Morgan) and RDKit 2D descriptors (labeled RDKit-2D) as molecule featurizers. Performance is assessed using mean absolute error on test split as the metric. ($\downarrow$ is better)}
\begin{tabular}{llcccc}
\hline
     & Descriptor           & Microsomal  & Microsomal  & Permeability    & Solubility      \\
     &                      & stability (human)              &  stability(rat)                &                 &                 \\
\hline
KAN  & Morgan   & $0.479 \pm 0.010$      & $0.559 \pm 0.012$      & $0.424 \pm 0.005$ & $0.476 \pm 0.006$ \\
UKAN & Morgan   & $0.476 \pm 0.003$      & $0.551 \pm 0.020$      & $\mathbf{0.403 \pm 0.006}$ & $\mathbf{0.435 \pm 0.009}$ \\
\hdashline
KAN  & RDKit-2D & $0.373 \pm 0.003$      & $0.492 \pm 0.010$      & $0.349 \pm 0.005$ & $0.386 \pm 0.002$ \\
UKAN & RDKit-2D & $0.368 \pm 0.008$      & $\mathbf{0.467 \pm 0.006}$      & $0.348 \pm 0.003$ & $0.397 \pm 0.013$ \\
\hline
\end{tabular}
\label{table:biogen_adme}
\end{table}

\section{conclusion}

In this work, we presented the Unbound Kolmogorov-Arnold Network (UKAN), which unifies multilayer perceptron networks (MLPs) with KANs along with an efficient GPU implementation of the underlying components of KANs. GPU acceleration decouples the computational cost and memory fingerprint of KANs from the grid size by using local matrix representations of B-spline functions. In addition, our proposed UKAN architecture allows the use of KANs without any fixed grid range limitation by generating coefficients from a coefficient-generator MLP. We evaluated UKAN model for regression, classification, and generative tasks. Our accompanying GPU library alleviates the core bottlenecks that have limited the practical scale of spline-based networks and serves as a reusable building block for future models. We expect UKAN and its variants to enable accelerated large-scale learning in domains such as molecular property prediction, protein docking, language, and vision, and we see promising directions in multi-GPU training, adaptive knot policies, and a deeper theory of approximation on unbounded domains.



\bibliography{iclr2026_conference}
\bibliographystyle{iclr2026_conference}

\appendix
\section{Appendix}
You may include other additional sections here.

\subsection{Cubic B-Spline Basis Matrix Representation}
\label{cubic_bspline}

For a B-spline of order 3, the basis matrix representation can be written as,
$$
    \operatorname{spline}(u) = \begin{bmatrix} 1 & u & u^2 & u^3 \end{bmatrix} \frac{1}{6} \begin{bmatrix}
1 & 4 & 1 & 0 \\
-3 & 0 & 3 & 0 \\
3 & -6 & 3 & 0 \\
-1 & 3 & -3 & 1
\end{bmatrix} \begin{bmatrix} c_0 \\ c_1 \\ c_2 \\ c_3 \end{bmatrix}.
$$

\subsection{UKAN mapping to KAN in Inference}
\label{KANUKAN}
\paragraph*{Proposition (UKAN $\rightarrow$ KAN at inference).}
Fix B\textendash spline degree $k$ and uniform grid spacing $h$, and set $K:=k+1$. Consider a UKAN whose coefficient\textendash generator (CG) produces a group vector $G_g\in\mathbb{R}^K$ for each group index $g\in\mathbb{Z}$. Evaluation uses the same local stencil as KAN,
\[
\text{spline}(x)=U(u)\,M_k\,(\cdot),\qquad
U(u)=(1,u,\ldots,u^k),\ \ u=\tfrac{x-t_i}{h}\in[0,1),
\]
with $t_i=i\,h$ the knot at cell index $i$. Let $\mathcal{X}\subset\mathbb{R}$ be the (finite) set of inputs on which inference is performed; more generally, let $\mathcal{I}$ be any compact interval containing $\mathcal{X}$. Then there exists a KAN with coefficients $\{C_i\}$ defined only for cells intersecting $\mathcal{I}$ such that the KAN exactly matches the UKAN on $\mathcal{X}$ (indeed, on all of $\mathcal{I}$).

\emph{Proof.}
For each cell index $i$ whose cell $[t_i,t_{i+1})$ intersects $\mathcal{I}$, write $i=gK+r$ with
\[
g=\Big\lfloor \tfrac{i}{K}\Big\rfloor,\qquad r=i\bmod K\in\{0,\ldots,K-1\}.
\]
Query the CG once (in inference mode; parameters fixed) to obtain the adjacent group vectors $G_{g-1},G_g\in\mathbb{R}^K$, and \emph{materialize} the KAN’s local coefficients by the sliding window
\[
C_i \;:=\; \mathrm{slice}\!\left(\,[G_{g-1};\,G_g],\; r:\,r+K\,\right)\in\mathbb{R}^K,
\]
where $[\,\cdot\,;\,\cdot\,]$ denotes concatenation and $\mathrm{slice}(\cdot,a{:}b)$ extracts entries $a,\ldots,b-1$.
By construction, for any $x$ in cell $i$ the UKAN evaluation applies the same linear stencil $U(u)M_k$ to the same length-$K$ window drawn from $[G_{g-1};G_g]$. Hence
\[
U(u)M_k\,C_i \;=\; U(u)M_k\,\mathrm{slice}\!\left(\,[G_{g-1};G_g],\,r{:}r+K\right),
\]
which is exactly the UKAN output in that cell. Because $\mathcal{I}$ intersects finitely many cells, the above defines finitely many $C_i$ and therefore a finite\textendash parameter KAN that agrees with the UKAN on all $x\in\mathcal{I}$, in particular on the finite inference set $\mathcal{X}$. 

\subsection{N-body Problem}
\label{n-body}
KANs promise better generalization compared to MLPs for regression tasks, similar to equivariant models allowing for the exploitation of symmetries for improved generalization. In particular, E(n)-Equivariant Graph Neural Networks (EGNNs) are equivariant with respect to the translations, rotations, and permutations \cite{satorras2022enequivariantgraphneural}. Here, we explore how combining equivariance with KAN leads to improved performance in the study of n-body systems as described in the EGNN paper \cite{satorras2022enequivariantgraphneural}. To evaluate this, we replace the final scalar predicting MLPs in EGNN with UKAN and KAN layers. Specifically, the scalar outputs of $\phi_x$ and $\phi_v$ in equations below are predicted with UKAN and KAN.

\begin{align*}
\mathbf{v}_i^{\ell+1}
&= \phi_v(\mathbf{h}_i^{\ell})\,\mathbf{v}_i^{\mathrm{init}}
   + C \sum_{j \ne i} (\mathbf{x}_i^{\ell} - \mathbf{x}_j^{\ell})\,\phi_x(\mathbf{m}_{ij}^{\ell+1}), \qquad
\mathbf{x}_i^{\ell+1} = \mathbf{x}_i^{\ell} + \mathbf{v}_i^{\ell+1}.
\end{align*}

Where $\mathbf{x}_i^{(\ell)}$ and $\mathbf{v}_i^{(\ell)}$ are the position and velocity of $i^{th}$ particle in the $l^{th}$ layer of EGNN. We keep the rest of parameters and datasets identical to the original paper and their code on Github. We also train the SE(3) Transformer model as another reference point. The results are shown in Table \ref{EGNN}, where we observe that UKAN and KAN improve the accuracy compared to the original architecture, and the improvement of UKAN is better than the KAN model.

\begin{table}[h]
\centering
\caption{Mean Squared Error for the future position prediction in the N-body system.}
\begin{tabular}{ll}
 \hline
 Method & MSE \\ 
 \hline
 EGNN & $0.00638$      \\ 
 EGNN+KAN  & $0.00609$  \\ 
 EGNN+UKAN & $\mathbf{0.00591}$ \\ 
 SE(3) Transformer & $0.02469$ \\ 
 \hline
\end{tabular}
\label{EGNN}

\end{table}

\subsection{Architecture Details of DDPM}
\label{model_arch}

The Decoder network is designed to transform input features through a series of linear and temporal layers. Here we explain architecture without KAN or UKAN layers\textit{, i.e.} with only linear layer and SiLU nonlinearity, and mention the differences at the end. The architecture consists of an input linear layer, three temporal layers, and an output linear layer. 

The Decoder is constructed with the following layers:

\begin{itemize}
    \item \textbf{Input Linear Layer:} The initial fully connected layer transforms the input features from the input dimension to an intermediate dimension.
    \item \textbf{Temporal Layers:} A series of temporal layers; specifically designed for the handling of time-dependent data. In our implementation, we use three temporal layers.
    \item \textbf{Output Linear Layer:} The final fully connected layer transforms the intermediate features back to the original input dimension.
    \item \textbf{Nonlinearity:} The intermediate features passed through SiLU non-linear activation function before being processed by the temporal layers.
\end{itemize}

The Temporal Layer is designed to integrate temporal information into the feature transformation process. This layer receives the input features and a temporal embedding, processes them through a series of linear transformations, and combines the outputs with a skip connection to ensure that the temporal information is effectively incorporated.

The Temporal Layer consists of the following components:

\begin{itemize}
    \item \textbf{Fully Connected Layers:} These layers perform linear transformations on the input features.
    \item \textbf{Temporal Encoding:} This layer projects the temporal embedding to the same dimensional space as the output features.
    \item \textbf{Skip Connection:} If the input and output features have the same dimension, an identity mapping is used. Otherwise, a linear transformation is applied to match the dimensions.
    \item \textbf{Output Linear Layer:} This layer produces the final output by combining the transformed features with the skip connection.
\end{itemize}

Within KAN and UKAN architectures, we only replaced the output linear layers of the Decoder network and Temporal layers with UKAN and KAN layers. We used UKAN with grid delta of 0.4 and 24 dimensional positional encoding and KAN with 11 grid between -2 and 2. Both KAN and UKAN were order 3 B-spline functions without MLP component.  

\subsection{Architecture details of molecular property prediction}
\label{mol_prop_pred_arch_appendix}
The model architecture consists of 3 layers of KAN/UKAN with hidden dimension equal to 2X the dimension of the input computed molecular features. The Morgan fingerprint was computed using RDKit package\cite{landrum2016rdkit} with a radius of 2 and a bit length of 1024, yielding a molecular features of dimension 1024. RDKit 2D descriptors were computed using RDKit package and normalized using descriptastorus package\cite{descriptastorus} resulting in a vector of size 200. Both KAN and UKANs were used with order 3 B-spline functions and a grid delta of $1.0$.

\subsection{Architecture Details of GCN}
\label{moleculenetappendix}
\paragraph{KAN/UKAN/FKAN-GCN} We build three GCN variants by replacing MLP block (node embedding, message passing, and readouts) with a KAN, UKAN, or FKAN layer, respectively. 

\textbf{Node embedding.}
Given initial node features $f_v \in \mathbb{R} ^ {d_{in}}$ and neighbor set $\mathcal{N}(v)$, we form an embedding by concatenating $f_v$ with a degree-normalized neighbor average and passing it through a basis layer $\Phi_E(\cdot)$ implemented by KAN/UKAN/FKAN:

$$
h_v^{(0)}  = \Phi_E \left( \big[f_v ;\tfrac{1}{|\mathcal{N}(v)|} \sum_{u \in \mathcal{N}(v) } f_u \big] \right) 
$$

\textbf{Message passing.}

At layer $\ell$, the neighbor is first transformed by the basis layer $\Phi_M^{\ell}(\cdot)$ and aggregated over the neighbor with summation. 

\[
m_{uv}^{(\ell)} \;=\;  \Phi_M^{(\ell)}\!\big(h_u^{(\ell)}\big),\qquad
h_v^{(\ell+1)} \;=\;\sum_{u\in\mathcal{N}(v)} m_{vu}^{(\ell+1)}.
\]

\textbf{Readout.}
After $L$ layers (four layers in our experiments), node features are pooled with \textsc{avg} (any permutation-invariant $P\in\{\textsc{avg},\textsc{sum},\textsc{max}\}$ is supported) and mapped to outputs by a KAN/UKAN/FKAN readout:
\[
\bar{h} \;=\; P\big(\{h_v^{(L)} : v\in V\}\big),\qquad
\hat{y} \;=\; \Phi_R(\bar{h}).
\]

Here, each $\Phi_{\{\!E,M,R\!\}}(\cdot)$ denotes the same structural layer instantiated with a different basis:
KAN uses fixed, bounded B-spline grids; UKAN uses a coefficient-generator to produce local B-spline coefficients on an unbounded symmetric grid; FKAN replaces the spline basis with a Fourier basis. All variants are drop-in compatible.

\section{Data prepreprocessing}
\subsection{Molecular property prediction}
\label{mol_prop_prep_data_appendix}
The molecular property prediction dataset was curated using standard practices in cheminformatics like removal of invalid SMILES, molecule standardization, removal salts, charge neutralization, and mean aggregation of any duplicate labels. This curation practice roughly following the guidelines outlined in \cite{fourches2010trust}. The molecules were split into train, validation, and test set using Bemis-Murcko scaffold split for each of the tasks in the dataset. In order to make statistically significant comparisons, only tasks containing a more than 500 labels were used. This resulted in selecting 4 out of the 6 tasks presented in the original paper.\cite{fang2023prospective}

\section{KAN/UKAN/FKAN-GCN results on MoleculeNet}
\label{moleculenet_results_appendix}
\cite{li2025kolmogorov} introduced a graph neural network with multi-layer perceptron layers replaced with KANs. The authors  demonstrate that such an architectures shows strong performance on the MoleculeNet datasets. However, we identified two issues with the KAN-GCN implementation by \cite{KangnnGithub}: (1) Best model selection is performed on test loss instead of validation loss, which is a deviation from the best practices in machine learning. (2) MoleculeNet consists of multi-task datasets with missing labels. The missing labels were assigned a value of $0.0$ instead of being treated as a missing label. Train and test metrics were computed based on this artificial label. We have fixed these issues in our reimplementation using warpKAN library.

To assess scalability of our implementation, we evaluate KAN, UKAN, and Fourier-based KAN (FKAN) graph convolutional neural network (GCN) on seven MoleculeNet datasets \cite{MoleculeNet}. The total number of components in the dataset is over 148,000 molecules, where each molecule  is composed of several heavy atoms, e.g., BACE and Tox21 have on average 65 and 36 atoms. We modify a standard GCN \cite{Wang_2023} by replacing its MLP blocks with KAN or UKAN layers (see Appendix \ref{moleculenetappendix}). To the best of our knowledge, our training is one of the large-scale training that incorporates KAN in all parts of GCN. Note while works like \cite{zhang2024graphkanenhancingfeatureextraction,bresson2025kagnnskolmogorovarnoldnetworksmeet} claim KAN usage it is usually limited to initial node embedding or readouts, as without efficient implementations like ours it is infeasible to investigate KAN in large scale. We train three models, where we divide data into training, validation, and testing datasets and select the best model based on the validation loss. We report the area under the curve (AUC) for models in Table \ref{table:moleculenet}. Each model is trained for 500 epochs using Adam optimizer with learning rate of 2e-4 and a step scheduler with a decay rate of 0.9 every 20 steps. Unlike values reported in \cite{li2025kolmogorov}, we see the accuracy of B-Spline- and Fourier-based models are very close to each other, when model selection is done correctly.

\begin{table}[h]
\centering
\caption{Performance of KAN, UKAN, and FKAN on MoleculeNet datasets ($\uparrow$ is better).}
\label{tab:moleculenet}
\begin{tabular}{l*{7}{c}}
\hline
Dataset & BBBP & BACE & ClinTox & Tox21 & SIDER & HIV & MUV \\
\hline
Tasks & 1 & 1 & 2 & 27 & 12 & 1 & 17 \\
\hline

KAN  & $65.6(2.2)$ & $80.0(2.4)$ & $96.5(0.2)$ & $79.4(0.7)$ & $83.5(0.2)$ & $76.5(1.3)$ & $71.8(5.7)$ \\
UKAN  & $64.5(2.2)$ & $74.9(6.3)$& $94.7(0.9)$ & $77.4(0.5)$ & $83.7(0.2)$ & $75.2(1.0)$ & $69.6(2.1)$ \\
FKAN & $67.0(2.0)$ & $80.5(3.0)$& $96.5(0.1)$ & $79.4(0.8)$ & $82.2(0.1)$ & $74.1(2.2)$ & $76.5(1.2)$ \\
\hline
\end{tabular}
\label{table:moleculenet}
\end{table}

\section{LLM Usage}

We used large-language models (LLMs) as general-purpose assistive tools during the preparation of this paper. Specifically:  

\begin{itemize}
    \item \textbf{Writing support:} LLMs were used to improve grammar, clarity, and text flow. All technical content, methodology, experimental design, and conclusions were conceived and verified by the authors.  
\item \textbf{Editing and formatting:} LLMs assisted in rephrasing sentences for readability, generating LaTeX table and figure formatting, and ensuring consistency in notation.  
\item \textbf{Brainstorming:} LLMs were used to explore alternative phrasings, organizational structures, and to verify the completeness of the literature-related sections.  
\end{itemize}
No LLM-generated text or ideas were included without careful review and verification by the authors. The models did not contribute to scientific novelty, research ideation, or experimental results. The authors assume full responsibility for all content in this manuscript.

\end{document}